\documentclass[table, 11pt]{article}
\usepackage[table, svgnames, dvipsnames]{xcolor}
\usepackage{tikz}
\usepackage[final]{acl}

\usepackage{times}
\usepackage{latexsym}
\usepackage{enumitem}
\usepackage[T1]{fontenc}
\usepackage[utf8]{inputenc}

\usepackage{microtype}

\usepackage{inconsolata}

\usepackage{graphicx}
\usepackage{amsmath}
\usepackage{amsfonts} 
\usepackage{bm}
\usepackage{multirow}
\usepackage{booktabs}
\usepackage{xspace}
\usepackage{adjustbox}
\DeclareMathOperator*{\argmin}{argmin}

\usepackage{colortbl}
\usepackage{subfigure}

\definecolor{lightgray}{gray}{0.80}
\definecolor{darkgreen}{rgb}{0.0, 0.5, 0.0}

\newcolumntype{g}{>{\columncolor{lightgray}}c}
\newcommand*{\method}{\textsc{COT$_2$Align}\@\xspace}

\title{\method: Cross-Chain of Thought Distillation via Optimal Transport Alignment for Language Models with Different Tokenizers}

\author{
  \textbf{Duc Anh Le\textsuperscript{1}\footnotemark[1]},
  \textbf{Tu Vu\textsuperscript{4}\footnotemark[1]},
  \textbf{Nam Le Hai\textsuperscript{1,\dag}\footnotemark[1]},
   \textbf{Diep Thi-Ngoc Nguyen\textsuperscript{2,3}},
  \\
   \textbf{Linh Ngo Van\textsuperscript{1}}, 
  \textbf{Trung Le\textsuperscript{5}},
  \textbf{Thien Huu Nguyen\textsuperscript{6}}
  \bigskip \\
\textsuperscript{1}Hanoi University of Science and Technology,
\textsuperscript{2}Oraichain Labs Inc., US, 
\textsuperscript{4}ByteDance Inc, \\
\textsuperscript{3}VNU University of Engineering and Technology,
\textsuperscript{5}Monash University,
\textsuperscript{6}University of Oregon
}

\begin{document}
\maketitle
\renewcommand{\thefootnote}{\fnsymbol{footnote}}
\footnotetext[1]{Equal contribution}
\footnotetext[2]{Corresponding author: \href{mailto:email@domain}{namlh@soict.hust.edu.vn}}
\renewcommand*{\thefootnote}{\arabic{footnote}}

\begin{abstract}
Large Language Models (LLMs) achieve state-of-the-art performance across various NLP tasks but face deployment challenges due to high computational costs and memory constraints. Knowledge distillation (KD) is a promising solution, transferring knowledge from large teacher models to smaller student models. However, existing KD methods often assume shared vocabularies and tokenizers, limiting their flexibility. While approaches like Universal Logit Distillation (ULD) and Dual-Space Knowledge Distillation (DSKD) address vocabulary mismatches, they overlook the critical \textbf{reasoning-aware distillation} aspect. To bridge this gap, we propose \method a universal KD framework that integrates Chain-of-Thought (CoT) augmentation and introduces Cross-CoT Alignment to enhance reasoning transfer. Additionally, we extend Optimal Transport beyond token-wise alignment to a sequence-level and layer-wise alignment approach that adapts to varying sequence lengths while preserving contextual integrity. Comprehensive experiments demonstrate that \method outperforms existing KD methods across different vocabulary settings, improving reasoning capabilities and robustness in domain-specific tasks.

% Large Language Models (LLMs) have demonstrated remarkable success across various NLP tasks. However, their real-world deployment remains challenging due to their substantial computational costs. Knowledge Distillation (KD) offers a promising solution by compressing LLMs into smaller, more efficient models. Yet, existing KD methods typically require the teacher and student models to share the same vocabulary and tokenizer, restricting architectural flexibility. Moreover, distilling LLMs with different vocabularies is rarely explored due to its inherent complexity. To address this challenge, we propose a novel approach that utilizes the Optimal Transport framework to align representations across different model layers. Additionally, we integrate Chain-of-Thought (COT) reasoning to enhance knowledge transfer. Extensive experiments across multiple benchmarks demonstrate that our method achieves state-of-the-art performance in LLM knowledge distillation.
\end{abstract}

\section{Introduction}\label{sec:intro}

Large Language Models (LLMs) have demonstrated remarkable capabilities across a wide range of natural language processing (NLP) tasks \citep{achiam2023gpt, touvron2023llama, jiang2023mistral, jiang2024mixtral, guo2025deepseek}. However, their deployment in real-world applications is often hindered by high computational costs, memory constraints, and latency issues. These limitations pose significant challenges to deploying LLMs efficiently on resource-constrained devices like mobile phones and IoT devices. Knowledge distillation (KD) \cite{hinton2015distilling} has emerged as a promising solution to this challenge by transferring knowledge from a large teacher model to a more compact student model, thereby retaining essential performance while reducing computational overhead. Conventional KD approaches generally seek to align the output distributions of teacher and student models through methods like Kullback-Leibler (KL) divergence \citep{zhang2023not, hsieh2023distilling, ko2024distillm}. 

For LLMs, knowledge distillation can be categorized into two main approaches: black-box KD and white-box KD. In black-box KD, the student model learns by mimicking the teacher model’s outputs, as it has no access to the teacher’s internal structure or variables \cite{fu2023specializing, kim2016sequence}. In contrast, white-box KD enables the student model to utilize the teacher model’s architecture and variables while constructing regularization constraints during training. Theoretically, this approach facilitates more comprehensive knowledge transfer, leading to superior performance \cite{wen2023f, gu2024minillm, ko2024distillm}. However, a fundamental limitation of these approaches is that they assume both models share the same vocabulary and tokenizer, a requirement that is increasingly impractical given the diversity of architectures and tokenization schemes used in contemporary LLMs. 

Several recent studies have attempted to address this issue by enabling KD across models with different tokenizers. For example, the Universal Logit Distillation (ULD) method \cite{boizard2024towards} utilizes Optimal Transport (OT) to align probability distributions at the token level across different vocabularies, offering a more flexible approach to distillation. Similarly, the Dual-Space Knowledge Distillation (DSKD) framework \citep{zhang2024dual} introduces a cross-model attention mechanism to unify output spaces, facilitating KD between models with non-overlapping token sets. Despite their advancements, these methods primarily emphasize a single aspect of knowledge transfer, overlooking a more comprehensive distillation process- \textbf{reasoning-aware distillation}. While LLMs are highly effective due to their advanced reasoning capabilities \citep{wei2022chain, huang2023towards, guo2025deepseek}, existing KD approaches for different vocabulary often overlook this critical component. Instead, they predominantly focus on aligning the final output, which may restrict the student model’s ability to improve reasoning skills and generalization capacity. Several KD methods for models with similar vocabularies have examined the effectiveness of CoT-augmented data in the distillation process \citep{ho2022large, hsieh2023distilling, ranaldi2024aligning}. However, these approaches integrate augmented data into student training without additional safeguards, increasing the risk of propagating flawed reasoning and hallucinations \citep{tonmoy2024comprehensive}, which may lead to final responses misaligned with the ground truth.

To overcome these challenges, \textbf{\textit{(1)}} we present a universal framework that improves distillation for models with different vocabularies by emphasizing additional aspects, particularly the teacher model’s reasoning ability. To achieve this, we integrate Chain-of-Thought (CoT) augmentation into the distillation process. Building on this, we introduce a novel Cross-CoT Alignment method that effectively transfers the teacher’s reasoning capability to the student model. Specifically, we design two alignment loss functions to encourage the student model to capture the teacher’s multi-step reasoning process: \textit{(a)} directly aligning the outputs of both models using the same input formats (standard and CoT) and \textit{(b)} aligning CoT and standard outputs to enhance the reliability of the CoT process in generating correct answers. However, aligning these outputs requires handling different vocabulary mappings. While existing KD methods for models with different vocabularies offer a potential proxy, we argue that these approaches are constrained by \textbf{token-wise alignment}. For instance, DSKD employs a simple linear projection to map distributions into a shared dimensional space for each token, which may overlook intricate relationships between teacher and student representations, leading to potential information loss. Similarly, ULD trims output token sequences to enforce token-wise alignment between teacher and student models, which can lead to incomplete sentences and misaligned tokens with varying contextual meanings, potentially distorting the semantic integrity of the distilled knowledge—especially when direct responses and CoT responses differ significantly in length. Therefore, \textbf{\textit{(2)}} we introduce a sequence-level alignment approach that operates without requiring projection into a uniform dimensional space or enforcing identical output lengths. Specifically, while we acknowledge OT as an effective solution for aligning the distinct distribution spaces of teacher and student models, as demonstrated in ULD, we extend its application beyond token-wise alignment. Instead, we propose \method, a sequence-level alignment method combined with a layer-by-layer approach to enhance reasoning knowledge transfer. This method effectively adapts to sequences of varying lengths, ensuring comprehensive output context alignment while preserving the integrity of the distilled knowledge. \textbf{\textit{(3)}} Our comprehensive experiments underscore the overall effectiveness of \method and the contribution of each proposed technique in enhancing existing universal KD methods. Additionally, when conducting domain-specific distillation across a wide range of tasks, we observe that the best-performing method, DSKD, exhibits limited superiority over other KD methods, contradicting the claims presented in their study. This finding suggests a lack of robustness in DSKD for domain-specific experiments, revealing gaps in its insights. In contrast, our approach demonstrates consistent improvements in this scenario, emphasizing its reliability and adaptability.

\section{Related Work and Background}
\subsection{Related Work}
\label{Related_work}
Knowledge Distillation (KD) for Large Language Models (LLMs) is categorized into black-box and white-box KD. Black-box KD relies solely on the teacher model's outputs \cite{fu2023specializing, kim2016sequence}, while white-box KD leverages the teacher's architecture, enabling more comprehensive knowledge transfer \cite{wen2023f, gu2024minillm, ko2024distillm}. White-box KD techniques enable alignment at different architectural levels, such as the teacher model's output distribution \cite{song2020lightpaff, liang2020mixkd}, hidden representations \cite{jiao2019tinybert, sun2019patient}, or attention scores \cite{wang2020minilm}. %For instance, TinyBERT \cite{jiao2019tinybert} aligns embeddings, hidden states, and self-attention layers, while MiniLLM \cite{zhang2023not} uses PTLoss to avoid imitating unreliable teacher outputs. Similarly, \cite{gu2024minillm} applies reverse KL divergence to generative models, reducing the risk of overestimating low-probability prediction

Although white-box KD has gained significant attention, most studies assume a shared tokenizer for simplicity, with limited exploration of KD across models with different vocabularies. This scenario poses challenges, as differing tokenization schemes lead to mismatched vocabulary sizes, complicating direct KL divergence loss computation. Solutions include MinED \cite{wan2024knowledge}, which uses dynamic programming to align logits by minimizing tokenized sequence edit distance; ULD \cite{boizard2024towards}, which employs a Wasserstein distance-based loss; and DSKD \cite{zhang2024dual}, which leverages Cross Model Attention (CMA) to align token representations in a shared space. DSKD is the current state-of-the-art for KD with mismatched vocabularies.

% Despite the growing interest in white-box KD, most research assumes a shared tokenizer between teacher and student models for simplicity. To the best of our knowledge, few studies have explored KD between models with different vocabularies. A major challenge in this setting is aligning the softmax outputs of the teacher and student models. Different tokenization schemes result in mismatched vocabulary sizes, making direct KL divergence loss computation infeasible. Several methods have been proposed to address this issue. MinED \cite{wan2024knowledge} employs dynamic programming to align logits by minimizing the edit distance between tokenized sequences. Universal Logit Distillation (ULD) \cite{boizard2024towards} replaces the standard KL divergence loss with a Wasserstein distance-based closed-form solution. Dual-space knowledge distillation (DSKD) \cite{zhang2024dual} introduces Cross Model Attention (CMA) to project token representations from different vocabularies into a shared space. Among these, DSKD has emerged as the state-of-the-art approach for LLM distillation across different vocabularies.

Additionally, Chain-of-Thought (CoT) prompting is a powerful technique in KD, as shown in studies like Fine-tune-CoT \cite{ho2022large}, which uses reasoning samples from large teachers to fine-tune smaller models, often surpassing the teacher's reasoning ability. Distilling step-by-step \cite{hsieh2023distilling} adds extracted rationales as supervision in a multi-task framework, enabling significant compression with less data. Instruction-tuning-CoT \cite{ranaldi2024aligning} guides students to generate structured reasoning, improving question-answering and mathematical tasks. Building on these approaches, we incorporate CoT to enhance knowledge transfer for student models.
% Additionally, Chain-of-Thought (CoT) prompting has proven to be a valuable technique in KD, as demonstrated in several studies \cite{ho2022large, hsieh2023distilling, chen2023mcc}. Fine-tune-CoT \cite{ho2022large} leverages reasoning samples from large teacher models to fine-tune smaller models, significantly enhancing their reasoning abilities—often surpassing the teacher model itself. Distilling step-by-step \cite{hsieh2023distilling} integrates extracted rationales as additional supervision in a multi-task learning framework, enabling substantial model compression while reducing training data requirements. Instruction-tuning-CoT \cite{ranaldi2024aligning} further refines CoT approach by guiding student models to generate structured, multi-step reasoning, improving their performance in question-answering and mathematical reasoning tasks. Inspired by these works, we incorporate the CoT mechanism into our method to enhance the knowledge transfer process for student models.
% \vspace{-0.5mm}
\subsection{Background}
\subsubsection{Knowledge Distillation}
Knowledge distillation \cite{hinton2015distilling} is a technique where a large, high-capacity model, often referred to as the teacher model, transfers its knowledge to a smaller, more efficient model, known as the student model. The goal is to train the student to mimic the teacher’s behavior, often by aligning the student’s predictions with the teacher's soft outputs (e.g., probability distributions). Given a vector of \textit{logit} $z$ as the outputs of the last fully connected layer of a deep model, the distillation loss can be formulated as:
\begin{equation}
    \mathcal{L}_{KD} (z_{t}, z_{s}),
\end{equation}
where $\mathcal{L}(.)$ indicates the divergence loss of logits, $z_{t}$ and $z_{s}$ are logits of the teacher and student, respectively. A typically chosen loss can be KL divergence. When training the student model, we also need a student loss of input and target. Note that, the student loss is always defined as the cross-entropy loss $\mathcal{L}_{CE}(y,p(z_{s}))$ between the ground truth label $y$ and the soft logit of the student model $p(z_{s})$. The final loss to train a student model is defined as:
\begin{equation}
    \mathcal{L} = \mathcal{L}_{CE}(y,p(z_{s})) + \mathcal{L}_{KD} (z_{t}, z_{s})
\end{equation}

% The conventional KD training paradigm is given in Fig. \ref{}. The teacher model is freezed in the distillation process while student models is trained. The Distillation Loss is computed after softmax layer with temperature $T=t$ to control the importance of each soft target. The Student Loss is computed after softmax layer with  temperature $T=1$ as usual.

% \begin{figure*}[htbp]
%     \centering
%     \includegraphics[width=1.0\textwidth]{imgs/hilton2015.png}
%     \caption{Knowledge Distillation Framework \citep{gou2021knowledge}.}
%     \label{fig:hilton2015}
% \end{figure*}

\subsubsection{Optimal Transport}

% Here's a paraphrased and more professional version of your Problem Formulation:

Optimal Transport (OT) \citep{villani2009optimal} is a mathematics framework to measure the dissimilarity between probability distributions. In comparison with others measures such as Kullback–Leibler divergence (KL) or Jensen-Shannon Divergence (JS) which require two distributions share the same support, OT does not require that condition. This feature enables OT in aligning different vocabularies of teacher and student models.

Formally, consider distributions are discrete. Given a complete separable metrics space $(\Omega,d)$, where $d:\Omega \times \Omega \to \mathbb{R}$ is the metrics on the space $\Omega$, let $P(\Omega)$ denote the set of all Borel probability measures on $\Omega$. Given to sets $\boldsymbol{X} = (\boldsymbol{x}_{1}, \boldsymbol{x}_{2}, ...\boldsymbol{x}_{N})$, $\boldsymbol{Y} = (\boldsymbol{y}_{1}, \boldsymbol{y}_{2}, ...\boldsymbol{y}_{M})$ of $N$ and $M$ sample points in $\Omega$, their empirical probability measures are defined as $f = \sum_{i=1}^{N} \alpha_{i} \delta_{\boldsymbol{x}_{i}} \in P(\Omega)$ 
and  $g = \sum_{j=1}^{M} \beta_{j} \delta_{\boldsymbol{y}_{j}} \in P(\Omega)$, respectively, where $\delta_{\boldsymbol{x}}$ is the Dirac unit mass on the position of $\boldsymbol{x}$ in $\Omega$, $\alpha_{i}$ and $\beta_{j}$ are the weight on the unit mass on $\boldsymbol{x}_i$, $\boldsymbol{y}_j$ respectively. Since $f$, $g$ are probability distributions, the weights vectors $\boldsymbol{\alpha} = (\alpha_1, \alpha_2, ... \alpha_N)$, $\boldsymbol{\beta} = (\beta_1, \beta_2, ... \beta_M)$ lie in the simplexes  $ \Theta_{N} := \{ \alpha_i \geq 0 \forall i = 1,...,N | \sum_{i=0}^{N} \alpha_i = 1 \}$ and $ \Theta_{M} := \{ \beta_j \geq 0 \forall j = 1,...,M | \sum_{i=0}^{M} \beta_j = 1 \}$
The empirical joint probability measure of $(\boldsymbol{X},\boldsymbol{Y})$ is denoted as: 
\begin{equation}
    h = \sum_{i=1}^{N} \sum_{j=1}^{M} \gamma_{ij}(\delta_{\boldsymbol{x_i}}, \delta_{\boldsymbol{y_j}})
\end{equation} whose marginal measures w.r.t $\boldsymbol{X}$ and $\boldsymbol{Y}$ are $f$ and $g$, respectively. The weight matrix $[\gamma_{ij}]_{ij}$ is a $N \times M$ non-negative matrix with row and column marginals $\boldsymbol{\alpha}$, $\boldsymbol{\beta}$. More concrete, $\sum_{i=1}^{N} \gamma_{ij} = \beta_{j} \forall j=1...M$ and $\sum_{j=1}^{M} \gamma_{ij} = \alpha_{i} \forall i=1...N$. The set of all the feasible weight matrixes is defined as the transportation polytope $U(\boldsymbol{\alpha}, \boldsymbol{\beta}$) of $\boldsymbol{\alpha}$, $\boldsymbol{\beta}$:
{\small\begin{equation}
    U(\boldsymbol{\alpha}, \boldsymbol{\beta}) := \{ \boldsymbol{T} \in \mathbb{R}_{+}^{N \times M} | \boldsymbol{T}\boldsymbol{1}_{M}=\boldsymbol{\alpha}, \boldsymbol{T}^T\boldsymbol{1}_{N}=\boldsymbol{\beta} \}.
\end{equation}}
An element $t_{ij}$ of a feasible $\boldsymbol{T}$ can be seen as the amount of mass transported from $\boldsymbol{x}_{i}$ to $\boldsymbol{y}_{j}$. The distance between $\boldsymbol{x}_{i}$ and $\boldsymbol{y}_{j}$ is measured by a metric $d$ raised to the power $p$. Matrix $\boldsymbol{D}$ is the pairwise distances between elements in $\boldsymbol{X}$ and $\boldsymbol{Y}$:
\begin{equation}
    \boldsymbol{D} := [d(\boldsymbol{x}_{i}, \boldsymbol{y}_{j})^p]_{ij} \in \mathbb{R}^{N \times M}.
\end{equation}
The cost of transporting $f$ to $g$ given a transport $\boldsymbol{T}$ is the Frobenius dot product between $\boldsymbol{T}$ and $\boldsymbol{D}$, which is $ \langle \boldsymbol{T},\boldsymbol{D} \rangle = tr(\boldsymbol{T}^T\boldsymbol{D})$ 

Given $\boldsymbol{\alpha}$, $\boldsymbol{\beta}$ and $\boldsymbol{D}$, the OT distance between empirical probability measures $f$ and $g$ is a linear programing problem:
\begin{equation}
    d_{W}(\boldsymbol{\alpha},\boldsymbol{\beta},\boldsymbol{D}) = \min_{\boldsymbol{T} \in U(\boldsymbol{\alpha},\boldsymbol{\beta})} \langle \boldsymbol{T},\boldsymbol{D} \rangle.
\end{equation}

Optimal Transport provides a framework to align two distribution with different supports. In the context of knowledge distillation LLMs with different tokenizers, Optimal Transport can be used to align distributions over two vocabularies. The distributions can be representations in each layers, or the final softmax of teacher and student models. These alignments can enhance the representations of student models in different layers, hence they can help the student model mimic the output of teacher.
\section{Proposed Method}

\subsection{Cross Chain-of-Thought Alignment}\label{sec:ccot}

Recent advances in Chain-of-Thought (CoT) reasoning have shown that multi-step reasoning can greatly enhance model performance \citep{wei2022chain, huang2023towards, feng2024towards}. However, current KD approaches \citep{zhang2024dual, wan2024knowledge, boizard2024towards} designed for models with different vocabularies mainly emphasize final output alignment and have not explored the ability to adequately capture and transfer intricate reasoning patterns from the teacher model. To bridge this gap, we not only integrate CoT augmentation into knowledge distillation for models with different vocabularies but also introduce a novel Cross-CoT Alignment method ($CCoT$). This approach aligns CoT-augmented samples with labeled responses within the \method framework (described in Section \ref{sec:otalign}), ensuring that both explicit outputs and the underlying reasoning processes are effectively transferred.

\paragraph{Data Augmentation with Chain-of-Thought:}  
The process begins by training or fine-tuning a teacher model on an initial dataset, resulting in a model with strong performance in a specific domain. To leverage its reasoning capabilities further, the teacher model is prompted with a zero-shot CoT instruction, such as \textit{“Let’s think step by step.”} This prompt encourages the model to generate a detailed, step-by-step rationale for its answers. 
Detail prompts can be found in Appendix \ref{sec:prompt}. The output typically includes two parts:  
\begin{itemize}
    \item The comprehensive reasoning process produced by the teacher model, referred to as \texttt{Teacher\_rationale}.
    \item A final answer in the format: [\texttt{Teacher\_rationale}]. \texttt{Therefore, the answer is: [Ground Truth]}.
\end{itemize}

These rationale-enriched outputs are then combined with the original dataset, creating a new CoT-augmented corpus that incorporates both direct responses and multi-step reasoning.

\paragraph{Cross-CoT Alignment:}
In order to achieve effective reasoning transfer, the student is trained on both CoT-augmented data and the teacher’s distilled outputs. This approach not only focuses on replicating the direct outputs but also on capturing the multi-step reasoning process that underlies them. To facilitate this, we consider the following pairs of representations:

\begin{itemize}[leftmargin=*]
    \item \((x, y)\) and \((x_{\text{CoT}}, y_{\text{CoT}})\) \\
    Here, \(x\) and \(x_{\text{CoT}}\) denote inputs without and with CoT prompts, respectively, while \(y\) and \(y_{\text{CoT}}\) are the corresponding student outputs. This pair focuses on transferring the reasoning embedded in the student's own responses.
    
    \item  \((x, Y)\) and \((x_{\text{CoT}}, Y_{\text{CoT}})\) \\
    In this pair, \(Y\) and \(Y_{\text{CoT}}\) represent the teacher’s outputs under standard and CoT conditions, respectively, emphasizing the direct transfer of the teacher’s reasoning.
\end{itemize}

Based on these pairs, we define two alignment losses aimed at reinforcing reasoning transfer:

\begin{enumerate}
    \item \textit{Cross Student-Teacher Output Alignment Loss}
    \begin{equation}\label{eq:cot_loss2}
        \mathcal{L}_{CST} = \mathcal{L}_{align}(\mathbf{y}_{\text{CoT}}, \mathbf{Y}_{\text{CoT}}) + \mathcal{L}_{align}(\mathbf{y}, \mathbf{Y}),
    \end{equation}
    
    \item \textit{Cross Standard-CoT Output Alignment Loss}
    \begin{equation}\label{eq:cot_loss1}
    \mathcal{L}_{CRC} = \mathcal{L}_{align}(\mathbf{y}, \mathbf{Y}_{\text{CoT}}) + \mathcal{L}_{align}(\mathbf{y}_{\text{CoT}}, \mathbf{Y})
\end{equation}
\end{enumerate}

Here, $\mathcal{L}_{align}$ refers to the alignment function, which will be detailed in Section \ref{sec:otalign}. The loss function (\ref{eq:cot_loss1}) aligns student and teacher outputs with identical input formats (standard and CoT), enabling the student to mimic the teacher’s behavior. Meanwhile, the loss function (\ref{eq:cot_loss2}) aligns CoT and standard outputs to enhance the reliability of the CoT process, reinforcing the transfer of accurate information and reasoning capabilities.

\subsection{Optimal Transport for Reasoning Distillation}\label{sec:otalign}

As discussed in Section \ref{sec:intro}, prior studies on knowledge distillation across different vocabularies have employed a stepwise approach, wherein the student model learns to mimic the teacher’s behavior by aligning its output logits with those of the teacher. While this method enables token-level distribution alignment, it imposes constraints such as requiring the same dimensional space for distributions or enforcing identical response lengths between the teacher and student, thereby limiting the full potential of reasoning knowledge transfer. To overcome these limitations, we introduce an Optimal Transport (OT)-based loss function, denoted as $\mathcal{L_{OT}}$, which facilitates sequence-level distribution alignment between the teacher and student models, effectively eliminating dependencies on sequence length and vocabulary differences.

% In conventional knowledge distillation \cite{hinton2015distilling}, the student model is trained to mimic the teacher by aligning its output logits with the teacher's predictions. However, this approach overlooks the rich intermediate representations captured by the teacher, limiting the effectiveness of the knowledge transfer. To address this, we propose an Optimal Transport (OT)-based loss, denoted as \(\mathcal{L}_{OT}\), which explicitly aligns the internal representations of the teacher and student models—even when their sequence lengths and vocabularies differ.

\paragraph{Empirical Distributions Definition} 
% Given tokenized sequences \(\mathbf{x} \in \mathbb{R}^{N \times d}\) and \(\mathbf{y} \in \mathbb{R}^{M \times D}\), we define their corresponding empirical distributions as:
% \begin{equation}
%     \mu = \sum_{i=1}^{N} \alpha_i \delta_{\mathbf{x}_i}, \quad \nu = \sum_{j=1}^{M} \beta_j \delta_{\mathbf{y}_j},
% \end{equation}
% where \(\alpha \in \Theta_N\) and \(\beta \in \Theta_M\) are probability vectors representing the mass distributions over tokens. Here, \(N\) and \(M\) denote the lengths of the student’s and teacher’s tokenized sequences, respectively.

Given that the distribution over tokens is uniform, we can simplify the empirical distributions as follows. For tokenized sequences \(\mathbf{x} \in \mathbb{R}^{N \times d}\) and \(\mathbf{y} \in \mathbb{R}^{M \times D}\), the empirical distributions are defined as:
\[
\mu = \frac{1}{N} \sum_{i=1}^{N} \delta_{\mathbf{x}_i}, \quad \nu = \frac{1}{M} \sum_{j=1}^{M} \delta_{\mathbf{y}_j},
\]
where each token in the student's sequence receives an equal mass of \(1/N\) and each token in the teacher's sequence receives an equal mass of \(1/M\). Here, \(N\) and \(M\) denote the lengths of the student’s and teacher’s tokenized sequences, respectively.

\paragraph{Cross-Attention Cost Matrix Computation} 
To align the sequences effectively, we construct a cost matrix \(\mathbf{C} \in \mathbb{R}^{N \times M}\) that quantifies the dissimilarity between token representations from the student and teacher. This is achieved in two main steps:
\begin{enumerate}
    \item \textbf{Similarity Matrix Computation:}  
    We compute the similarity matrix:
    \begin{equation}
        \mathbf{S} = \frac{\mathbf{X} \, \mathbf{P}(\mathbf{Y})^\top}{\sqrt{d}},
    \end{equation}
    where \(\mathbf{X} \in \mathbb{R}^{N \times d}\) and \(\mathbf{Y} \in \mathbb{R}^{M \times D}\) are the token embeddings for the student and teacher, respectively. The mapping matrix \(\mathbf{P} \in \mathbb{R}^{D \times d}\) projects the teacher’s embeddings into the student’s space, and the scaling factor \(\sqrt{d}\) ensures numerical stability.
    
    \item \textbf{Normalization and Cost Computation:}  
    The similarity matrix is normalized row-wise using the softmax function:
    \begin{equation}
        \mathbf{S}_{\text{norm}} = \mathrm{softmax}(\mathbf{S}),
    \end{equation}
    ensuring that each row sums to 1. The cost matrix is then derived as:
    \begin{equation}
        \mathbf{C} = \mathbf{1} - \mathbf{S}_{\text{norm}},
    \end{equation}
    where \(\mathbf{1}\) denotes a matrix of ones.
\end{enumerate}

\paragraph{Optimal Transport Plan Computation} 
We compute the optimal transport plan \(\mathbf{T}^*\) by solving the entropy-regularized OT problem \cite{distances2013lightspeed}:
\begin{equation}
    \mathbf{T}^* = \arg\min_{\mathbf{T} \in U(\alpha, \beta)} \langle \mathbf{T}, \mathbf{C} \rangle - \frac{1}{\lambda} H(\mathbf{T}),
\end{equation}
where \(H(\mathbf{T}) = - \sum_{i,j} T_{ij} \log T_{ij}\) is the entropy regularization term. The OT distance is then defined as:
\begin{equation}
    \mathcal{L}_{OT}(\mathbf{x}, \mathbf{y}) = \langle \mathbf{T}^*, \mathbf{C} \rangle.
\end{equation}
This loss enforces structural alignment between the teacher and student representations.

\paragraph{Layer-wise Optimal Transport Alignment} 
To ensure comprehensive knowledge transfer, we apply the OT distance to both the embedding and last hidden layers:
{\small\begin{equation}
    \mathcal{L}_{\mathcal{OT}}(s, t) = \mathcal{L}_{OT}\bigl(e^s_{1:N}, \, e^t_{1:M}\bigr) + \mathcal{L}_{OT}\bigl(h^s_{1:N}, \, h^t_{1:M}\bigr),
\end{equation}}
where \(e^s\) and \(e^t\) denote the student and teacher embeddings, and \(h^s\) and \(h^t\) represent their respective last hidden states.
The loss function $\mathcal{L}_{\mathcal{OT}}$ will serve as $\mathcal{L}_{align}$ in Equations (\ref{eq:cot_loss1}) and (\ref{eq:cot_loss2}).

\subsection{Overall Knowledge Distillation Objective}

\method extends DSKD \cite{zhang2024dual} by incorporating the techniques introduced in Sections \ref{sec:ccot} and \ref{sec:otalign}. Thus, our final knowledge distillation objective is formulated as follows:
\begin{equation}
    \mathcal{L}_{CCoT} = \mathcal{L}_{CRC} + \mathcal{L}_{CST},
\end{equation}
\begin{equation}\label{eq:final_obj}
    \mathcal{L} = (1 - \alpha) \, \mathcal{L}_{CE} + \alpha \left(\mathcal{L}_{CCoT} + \mathcal{L}_{KD}\right),
\end{equation}
where \(\alpha \in [0,1]\) balances the contributions of the traditional cross-entropy loss \(\mathcal{L}_{CE}\), the original DSKD's distillation loss \(\mathcal{L}_{KD}\), and the proposed OT-based loss $\mathcal{L}_{CCoT}$. This comprehensive objective enables the student model not only to replicate the teacher's final output at the token-wise level but also to align at the sequence level, capturing the underlying reasoning process; thereby ensuring a more effective knowledge transfer process.

\section{Experiments}

\begin{table*}[t]
\footnotesize
\centering
\begin{adjustbox}{width=\textwidth}
\begin{tabular}{l|c|c|c|c|c}
\toprule
\textbf{Methods} & \textbf{Dolly} & \textbf{Alpaca} & \textbf{S-NI} & \textbf{Dialogue Sum} & \textbf{Avg} \\ 
\bottomrule
\toprule
\multicolumn{6}{c}{\textit{\textbf{Qwen1.5-1.8B $\rightarrow$ GPT2-120M}}} \\ 
\hline
% \multicolumn{6}{c}{\textit{Teacher: GPT2-1.5B (Same Vocabulary)}} \\ 
% \hline
% Teacher 
% & $26.68_{\pm 0.23}$	
% & $30.01_{\pm 0.36}$	
% & $24.57_{\pm 0.21}$  
% & $32.00_{\pm 0.27}$
% & $28.32$ \\  
% \cline{1-6}
% KL 
% & $24.20_{\pm 0.37}$	
% & $27.55_{\pm 0.59}$	 
% & $21.32_{\pm 0.09}$	 
% & $31.29_{\pm 0.30}$  
% & $26.09$ \\  
% \textbf{KL + \method} 
% & $\mathbf{24.80_{\pm 0.32}}$	
% & $\mathbf{28.14_{\pm 0.30}}$	
% & $\mathbf{21.85_{\pm 0.12}}$	 
% & $\mathbf{31.36_{\pm 0.09}}$  
% & $\mathbf{26.54}_{\color{darkgreen}\uparrow 0.45}$ \\  
% \hline  
% \multicolumn{6}{c}{\textit{Teacher: Qwen1.5-1.8B (Different Vocabularies)}} \\
% \hline
Teacher 
& $28.23_{\pm 0.13}$	
& $33.76_{\pm 0.33}$	
& $30.32_{\pm 0.26}$	
& $35.37_{\pm 0.26}$ 
& $31.92$ \\  
\cline{1-6}
SFT 
% 23.78 +- 0.38		27.2 +- 0.23			20.45 +- 0.23		30.25 +- 0.22
& $23.78_{\pm 0.38}$	
& $27.20_{\pm 0.23}$	
& $20.45_{\pm 0.23}$  
& $30.25_{\pm 0.22}$
& $25.42$ \\  
ULD \cite{boizard2024towards}
& $23.77_{\pm 0.41}$	
& $27.50_{\pm 0.50}$	
& $21.37_{\pm 0.34}$	
& $30.23_{\pm 0.10}$ 
& $25.72$ \\ 
% \textbf{ULD + \method} 
% & $\mathbf{23.83_{\pm 0.28}}$	
% & $\mathbf{28.12_{\pm 0.41}}$	
% & $\mathbf{21.76_{\pm 0.16}}$	 
% & $\mathbf{30.43_{\pm 0.12}}$ 
% & $\mathbf{26.04}_{\color{darkgreen}\uparrow 0.32}$ \\ 
MinED \cite{wan2024knowledge}
& $24.21_{\pm 0.31}$	
& $28.47_{\pm 0.42}$	
& $21.76_{\pm 0.13}$	 
& $31.36_{\pm 0.09}$ 
& $26.45$ \\ 
% \textbf{MinED + \method} 
% & $\mathbf{24.63_{\pm 0.36}}$	
% & $\mathbf{28.86_{\pm 0.44}}$	
% & $\mathbf{22.18_{\pm 0.19}}$	 
% & $\mathbf{31.58_{\pm 0.24}}$ 
% & $\mathbf{26.81}_{\color{darkgreen}\uparrow 0.36}$ \\ 
DSKD \cite{zhang2024dual}
& $24.42_{\pm 0.32}$	
& $28.48_{\pm 0.32}$    
& $22.26_{\pm 0.26}$ 
& $31.46_{\pm 0.22}$ 
& $26.66$ \\ 
\textbf{\method} 
& $\mathbf{25.34_{\pm 0.23}}$  
& $\mathbf{28.78_{\pm 0.18}}$	
& $\mathbf{24.02_{\pm 0.32}}$	
& $\mathbf{31.76_{\pm 0.10}}$ 
& $\mathbf{27.48}_{\color{darkgreen}\uparrow 0.82}$ \\  
\bottomrule
\toprule
\multicolumn{6}{c}{\textit{\textbf{Mistral-7B $\rightarrow$  TinyLLaMA-1.1B}}} \\ 
\hline
% \multicolumn{6}{c}{\textit{Teacher: Llama2-7B (Same Vocabulary)}} \\ 
% \hline Teacher
% & $32.15_{\pm 0.56}$ 
% & $36.44_{\pm 0.48}$ 
% & $30.16_{\pm 0.25}$ 
% & $36.18_{\pm 0.23}$ 
% & $33.73$ \\  
% \cline{1-6}
% KL
% & $25.75_{\pm 0.26}$ 
% & $31.83_{\pm 0.57}$ 
% & $26.73_{\pm 0.17}$ 
% & $32.83_{\pm 0.18}$ 
% & $29.29$ \\  
% \textbf{KL + \method} 
% & $\mathbf{26.23_{\pm 0.31}}$ 
% & $\mathbf{32.34_{\pm 0.39}}$ 
% & $\mathbf{27.97_{\pm 0.28}}$ 
% & $\mathbf{33.42_{\pm 0.11}}$ 
% & $\mathbf{29.99}_{\color{darkgreen}\uparrow 0.7}$  \\  
% \hline
% \multicolumn{6}{c}{\textit{Teacher: Mistral-7B (Different Vocabularies)}} \\ 
% \hline 
Teacher 
% 32.15 +- 0.56		36.44 +- 0.48			30.16 +- 0.25		36.18 +- 0.23

& $32.15_{\pm 0.56}$ 
& $36.44_{\pm 0.48}$ 
& $30.16_{\pm 0.25}$ 
& $36.18_{\pm 0.23}$ 
& $33.73$ \\ 
\cline{1-6}
SFT 
& $23.20_{\pm 0.16}$ 
& $29.48_{\pm 0.48}$ 
& $24.65_{\pm 0.25}$ 
& $31.08_{\pm 0.17}$ 
& $27.10$ \\
ULD \cite{boizard2024towards}
& $25.48_{\pm 0.29}$ 
& $31.33_{\pm 0.36}$ 
& $26.55_{\pm 0.10}$ 
& $33.69_{\pm 0.26}$ 
& $29.26$ \\ 
% \textbf{ULD + \method} 
% & $\mathbf{25.63_{\pm 0.22}}$ 
% & $\mathbf{31.86_{\pm 0.53}}$ 
% & $\mathbf{27.72_{\pm 0.33}}$ 
% & $\mathbf{33.90_{\pm 0.14}}$ 
% & $\mathbf{29.78}_{\color{darkgreen}\uparrow 0.52}$ \\ 
MinED \cite{wan2024knowledge}
& $25.54_{\pm 0.59}$ 
& $31.82_{\pm 0.33}$ 
& $26.13_{\pm 0.23}$ 
& $33.31_{\pm 0.16}$ 
& $29.20$ \\ 
% \textbf{MinED + \method} 
% & $\mathbf{26.32_{\pm 0.52}}$ 
% & $\mathbf{32.76_{\pm 0.27}}$ 
% & $\mathbf{28.13_{\pm 0.13}}$ 
% & $\mathbf{33.42_{\pm 0.20}}$ 
% & $\mathbf{30.16}_{\color{darkgreen}\uparrow 0.96}$ \\
DSKD \cite{zhang2024dual}
& $26.28_{\pm 0.35}$ 
& $32.31_{\pm 0.15}$ 
& $26.74_{\pm 0.24}$ 
& $33.44_{\pm 0.18}$ 
& $29.69$ \\ 
\textbf{\method} 
& $\mathbf{27.41_{\pm 0.43}}$  
& $\mathbf{33.31_{\pm 0.49}}$ 
& $\mathbf{29.77_{\pm 0.20}}$  
& $\mathbf{35.01_{\pm 0.20}}$ 
& $\mathbf{31.38}_{\color{darkgreen}\uparrow 1.69}$ \\ 
\bottomrule   
\toprule 
\multicolumn{6}{c}{\textit{\textbf{Qwen2.5-7B-Instruct $\rightarrow$ GPT2-1.5B}}} \\  
\hline
Teacher 
& $28.49_{\pm 0.21}$	
& $35.75_{\pm 0.25}$	
& $32.35_{\pm 0.24}$	
& $35.24_{\pm 0.08}$ 
& $32.96$ \\  
\cline{1-6}
SFT
& $21.83_{\pm 0.28}$	
& $27.15_{\pm 0.31}$	
& $23.16_{\pm 0.15}$	
& $30.74_{\pm 0.17}$ 
& $25.72$ \\
ULD \cite{boizard2024towards}
& $24.52_{\pm 0.28}$	
& $29.17_{\pm 0.22}$	
& $24.18_{\pm 0.08}$	
& $32.74_{\pm 0.35}$ 
& $27.65$ \\ 
% \textbf{ULD + \method} 
% & $\mathbf{24.73_{\pm 0.33}}$	
% & $\mathbf{30.22_{\pm 0.18}}$	
% & $\mathbf{25.07_{\pm 0.23}}$	
% & $\mathbf{33.30_{\pm 0.42}}$ 
% & $\mathbf{28.33}_{\color{darkgreen}\uparrow 0.68}$ \\ 

MinED \cite{wan2024knowledge}
& $25.52_{\pm 0.44}$	
& $30.41_{\pm 0.56}$	
& $25.09_{\pm 0.25}$	
& $33.83_{\pm 0.24}$	         
& $28.71$ \\ 
% \textbf{MinED + \method} 
% & $\mathbf{25.85_{\pm 0.28}}$  
% & $\mathbf{30.43_{\pm 0.28}}$	
% & $\mathbf{25.74_{\pm 0.22}}$	
% & $\mathbf{34.79_{\pm 0.11}}$ 
% & $\mathbf{29.20}_{\color{darkgreen}\uparrow 0.49}$ \\ 

DSKD \cite{zhang2024dual}
& $25.38_{\pm 0.46}$	           
& $30.48_{\pm 0.38}$  
& $25.92_{\pm 0.18}$	 
& $33.82_{\pm 0.23}$            
& $28.90$ \\ 
\textbf{\method} 
& $\mathbf{26.72_{\pm 0.22}}$          
& $\mathbf{33.02_{\pm 0.40}}$	      
& $\mathbf{27.72_{\pm 0.13}}$	     
& $\mathbf{35.63_{\pm 0.22}}$              
& $\mathbf{30.77}_{\color{darkgreen}\uparrow 1.87}$ \\ 
\bottomrule
\end{tabular}
\end{adjustbox}
\caption{Comparison of methods across different datasets. We present the $mean_{\pm std}$ values derived from experiments conducted across 5 random seeds. SFT refers to Supervised Fine-Tuning, where the student model is directly trained on the downstream dataset.}
\label{tab:main_result}
\end{table*}

\begin{table}[ht]
\centering
\begin{adjustbox}{width=0.49\textwidth}
\begin{tabular}{l|c|c|c|c}
\toprule
\textbf{Methods} & \textbf{KL} & \textbf{ULD} & \textbf{MinED} & \textbf{DSKD} \\
\bottomrule
\toprule
\multicolumn{5}{c}{\textit{\textbf{GPT2-1.5B$^\dag$ \& Qwen1.5-1.8B$^\ddag$ $\rightarrow$ GPT2-120M}}} \\ 
\hline
Original & $26.09$ & $25.72$ & $26.45$ & $26.66$ \\
\quad + \textbf{ours} &
$\mathbf{26.54}_{\color{darkgreen}\uparrow 0.12}$ & $\mathbf{26.04}_{\color{darkgreen}\uparrow 0.32}$ & $\mathbf{26.81}_{\color{darkgreen}\uparrow 0.36}$ & $\mathbf{27.48}_{\color{darkgreen}\uparrow 0.82}$ \\
\toprule
\multicolumn{5}{c}{\textit{\textbf{Llama2-7B$^\dag$ \& Mistral-7B$^\ddag$ $\rightarrow$ TinyLLaMA-1.1B}}} \\ 
\hline
Original & $29.29$ & $29.26$ & $29.20$ &  $29.69$ \\
\quad + \textbf{ours} & $\mathbf{29.99}_{\color{darkgreen}\uparrow 0.7}$ & $\mathbf{29.78}_{\color{darkgreen}\uparrow 0.52}$ & $\mathbf{30.16}_{\color{darkgreen}\uparrow 0.96}$ & $\mathbf{31.38}_{\color{darkgreen}\uparrow 1.69}$\\
 
\toprule 
\multicolumn{5}{c}{\textit{\textbf{Qwen2.5-7B-Instruct$^\ddag$ $\rightarrow$ GPT2-1.5B}}} \\  
\hline
Original & - & $27.65$ & $28.71$ & $28.90$ \\
\quad + \textbf{ours} & - & $\mathbf{28.33}_{\color{darkgreen}\uparrow 0.68}$ & $\mathbf{29.20}_{\color{darkgreen}\uparrow 0.49}$ & $\mathbf{30.77}_{\color{darkgreen}\uparrow 1.87}$ \\
\hline
\end{tabular}
\end{adjustbox}
\caption{Performance evaluation of our method across a diverse range of baseline models. We report the average ROUGE score across four tasks. Here, KL represents the method designed for similar vocabulary scenarios. while $\dag$ indicates teacher models used in the similar vocabulary setting, $\ddag$ denotes teacher models applied in the different vocabulary setting. DSKD + ours is \method.}
\label{tab:universal_rs}
\end{table}

% In this section, we provide experimental results and analysis to demonstrate the effectiveness of our proposed \method.

\subsection{Experimental Setup}\label{sec:exp_setup}

In the state-of-the-art method DSKD \cite{zhang2024dual}, the authors typically use a single dataset for distillation and assess performance across multiple datasets covering different domains or tasks during testing. Although the reported results are based on the best checkpoint across various tasks, Figure \ref{fig:scenario} indicates that this checkpoint may not accurately reflect the effectiveness of the distillation process on the in-domain training dataset. Specifically, we independently create training/validation/testing sets for each domain to enhance knowledge distillation within a specific domain. The datasets are specifically detailed as follows:

\paragraph{Data.} To conduct the knowledge distillation process, we choose four datasets \textsc{databricks-dolly-15k} (\textbf{Dolly}) processed by \cite{gu2024minillm}; \textsc{alpaca} (\textbf{Alpaca}) \cite{alpaca}; \textsc{S-NI} (\textbf{S-NI}) \cite{wang-etal-2022-super}; and \textsc{dialogsum} (\textbf{Dialogsum}) \cite{chen2021dialogsum}. For Alpaca, we retain only samples with response lengths in \([11, +\infty)\) and split them into training, validation, and test sets. Since the original S-NI dataset provides only training and test splits, we further partition its training set, selecting samples with lengths in \([6, +\infty)\) to create separate training and development subsets, and we filter the test set to include only samples with lengths in \([11, +\infty)\). Detailed statistics for each dataset are provided in Table \ref{tab:dataset}.

\paragraph{Baselines.} We apply our method and compare it against three state-of-the-art baselines, \textbf{ULD} \cite{boizard2024towards}, \textbf{MinED} \cite{boizard2024towards}, and \textbf{DSKD} \cite{zhang2024dual}, which utilize KD techniques on models with different vocabularies. Detailed descriptions of these baselines are provided in Section \ref{Related_work}.

Further details on the models used, as well as the training and evaluation setup, can be found in Appendix \ref{sec:appendix_exp}.

\subsection{Main Results}

\begin{figure}[t]
    \centering
    \adjustbox{max width=0.49\textwidth}{
        \includegraphics{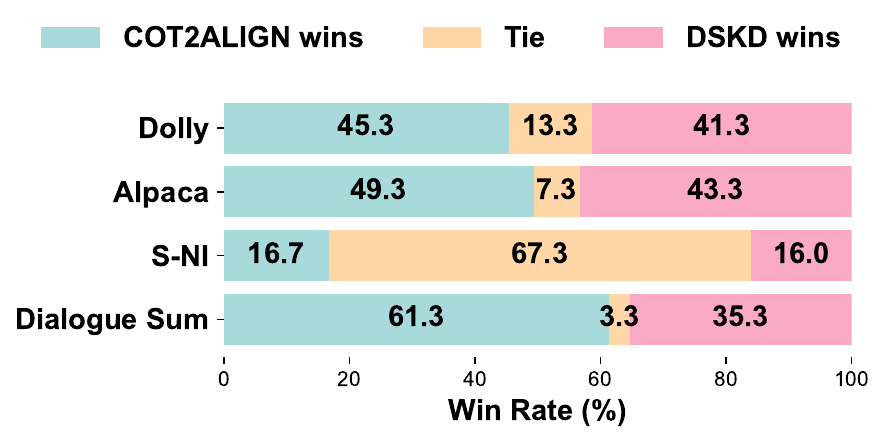}
    }
    \caption{Win rate comparison across categories for DSKD and \method from Qwen2.5-7B-Instruct to GPT2-1.5B}
    \label{fig:win_rate_comparison}
\end{figure}

Tables \ref{tab:main_result} and \ref{tab:universal_rs} present the performance across various methods and datasets. Overall, our proposed approach consistently outperforms all baselines across various scenarios, demonstrating its effectiveness in diverse settings.

% \paragraph{Same Vocabulary Scenario:} Although our method is designed for models with different vocabularies, it can be universally applied, including scenarios where the teacher and student share the same vocabulary. In this case, we evaluate the effectiveness of \method by comparing it with a simple KD approach utilizing KL loss. The results indicate that applying our method achieves an improvement of 0.4–0.7\% in settings using TinyLLama-1.1B and GPT-120M, demonstrating its effectiveness in this scenario.

\paragraph{\method vs. State-of-the-Art Baselines:} Table \ref{tab:main_result} compares \method with baseline methods in the different vocabulary scenario, demonstrating that our approach significantly improves the performance of state-of-the-art baselines. Specifically, compared to the strongest baseline, DSKD, achieves nearly a 2\% improvement when using TinyLLama-1.1B and GPT-1.5B, and a 0.82\% gain when distilling models with over 1B parameters into GPT-120M. In addition to match-based metrics (e.g., ROUGE), we evaluate other qualitative aspects of the student model's responses, including helpfulness, relevance, accuracy, depth, and creativity, using the API of gpt4-turbo-0409 as the judge. We follow the prompt presented by \citet{zhang2024dual} and present in Figure \ref{fig:prompts}. The results, illustrated in Figure \ref{fig:win_rate_comparison}, reveal that after applying our approach to DSKD, the instances where the judge determines our responses to win or tie significantly exceed the loss rate on all 4 benchmarks. This highlights the improved naturalness and correctness achieved through \method, particularly by emphasizing the aspect of reasoning distillation.

\paragraph{Universal Applicability of Our Framework:} While our method is based on DSKD, it can be universally applied to any KD approach by substituting the $\mathcal{L}_{KD}$ term in Equation (\ref{eq:final_obj}) with the original KD loss of the respective method. Table \ref{tab:universal_rs} presents the results for this experiment on diverse KD baselines and models. Table \ref{tab:universal_rs} presents the results of this experiment across diverse KD baselines and models. The findings indicate consistent improvements with our framework, demonstrating its effectiveness in enhancing knowledge distillation performance across various methods and models in both similar and different vocabulary scenarios.

\subsection{Analysis}

\begin{figure}[t]
\centering
\begin{adjustbox}{width=0.48\textwidth}
\includegraphics{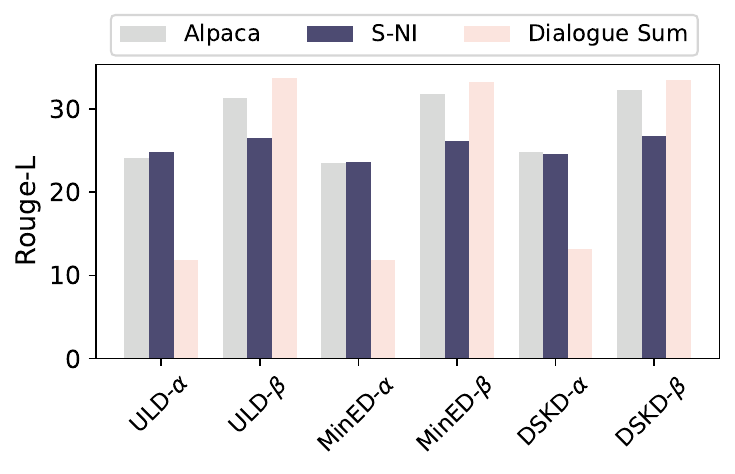}
\end{adjustbox}
\caption{Comparison of performance across various methods under the DSKD setting ($\alpha$) and domain-specific setting ($\beta$).}
\label{fig:scenario}
\end{figure}

\paragraph{Domain-Specific Scenarios Expose DSKD Limitations:} Unlike prior experimental setups of DSKD, we conduct both training and evaluation of the distillation process on domain-specific datasets, allowing the models to be fully optimized for specific applications on edge devices (as discussed in Section \ref{sec:exp_setup}). Figure \ref{fig:scenario} demonstrates that applying this scenario enables the model to achieve significantly higher performance on specific tasks (up to 20\%) compared to training under the previous settings. This approach enables a more in-depth exploration of distillation on model states that have been optimized for distinct task-specific objectives. Moreover, the results indicate that under our scenario, previous baselines do not demonstrate substantial dominance over others. For instance, in Table 1, DSDK outperforms MinED by only marginal proportions of 0.2\% when distilling Qwen1.5-1.8B to GPT-120M and Qwen2.5-7B-Instruct to GPT-1.5B. While DSDK reports significant performance (over 6\%) compared to MinED using their settings, this might not indicate the true effectiveness of each method among others on specific domain evaluation. Conversely, our method consistently enhances performance across various techniques, showcasing its effectiveness and reliability in domain-specific scenarios.

\paragraph{Adaptability Across Diverse Scales:} In our experiments, we utilize two student model sizes: super small models (approximately 100M parameters, such as GPT2-120M) and small models (around 1B parameters, including TinyLLaMA-1.1B and GPT2-1.5B), paired with 1B and 7B teacher models, respectively. Across these varying scales, our method consistently demonstrates its effectiveness in enhancing existing KD techniques, showcasing its adaptability and scalability for practical use. Notably, we observe more substantial improvements when employing larger student and teacher models. This can be attributed to smaller teachers inherently exhibiting weaker reasoning capabilities \citep{shridhar2022distilling, bi-etal-2025-enhancing}, which results in less reliable responses and reduces the impact of the reasoning-aware mechanism in \method. Specifically, experiments with larger student-teacher pairs reveal that instruction-tuned teachers (e.g., Qwen 2.5 Instruct) offer greater benefits compared to base pretrained teacher models (i.e. LLaMa and Mistral). These findings underscore the significance of reasoning-focused distillation and highlight the effectiveness of ours approach.

\paragraph{Ablation Study:} Table \ref{tab:methods_ablation} provides an ablation study that highlights the individual contributions of each proposed component within \method.  Specifically, DSKD + \method achieves the highest scores on all datasets, demonstrating its superiority in the knowledge distillation process (e.g., improvements on Dolly from 26.28 to 27.41 and on Dialogue Sum from 33.44 to 35.01). Both proposed losses individually enhance baseline performance (from 0.6 to over 1\%), underscoring their effectiveness. Notably, $\mathcal{L}_{CRC}$ demonstrates greater improvement, highlighting the effectiveness of our constraint in guiding the CoT response to be more reliable. We also conduct experiments using CoT augmentation into the original DSKD. The expermiental results are presented in the line "DSKD + only COT".  It is evident that while CoT augmentation improves DSKD, its impact is not as significant as \method.

 Moreover, even when applied solely to the last hidden layers, the improvement remains evident, further validating the effectiveness of our approach. Finally, the combination of these components in \method yields best performance, highlighting their complementary nature and the ability to collectively optimize knowledge transfer. These results confirm the robustness of the proposed approach in enhancing the baseline KD method.

\begin{table}[t]
\centering
\begin{adjustbox}{width=0.49\textwidth}
\begin{tabular}{l|c|c|c|c}
\toprule
\textbf{Methods}  & \textbf{Dolly} & \textbf{Alpaca} & \textbf{S-NI} & \textbf{Dialogue Sum}             \\ 
\midrule 
\multicolumn{5}{c}{\textbf{Mistral $\to$ TinyLLaMA}} \\ 
\hline
DSKD                     & 26.28\textsubscript{$\pm$0.35}     & 32.31\textsubscript{$\pm$0.15}     & 26.74\textsubscript{$\pm$0.24}     & 33.44\textsubscript{$\pm$0.18}     \\ 

\quad + only COT               & 27.27\textsubscript{$\pm$0.44}     & 32.42\textsubscript{$\pm$0.24}     & 28.97\textsubscript{$\pm$0.27}     & 33.74\textsubscript{$\pm$0.16}     \\ 

\quad + $\mathcal{L}_{CST}$                & 26.93\textsubscript{$\pm$0.40}     & 32.99\textsubscript{$\pm$0.41}     & 28.07\textsubscript{$\pm$0.16}     & 34.47\textsubscript{$\pm$0.20}     \\ 

\quad + $\mathcal{L}_{CRC}$               & 26.93\textsubscript{$\pm$0.46}     & 33.18\textsubscript{$\pm$0.58}     & 29.51\textsubscript{$\pm$0.24}     & 34.39\textsubscript{$\pm$0.24}     \\ 
\quad + only hidden state layers              & 27.07\textsubscript{$\pm$0.56}     & 33.10\textsubscript{$\pm$0.23}     & 29.41\textsubscript{$\pm$0.14}     & 34.80\textsubscript{$\pm$0.20}     \\ 
\textbf{\method}    & \textbf{27.41\textsubscript{$\pm$0.43}} & \textbf{33.31\textsubscript{$\pm$0.49}} & \textbf{29.77\textsubscript{$\pm$0.20}} & \textbf{35.01\textsubscript{$\pm$0.20}} \\ 
\bottomrule
\end{tabular}
    
\end{adjustbox}
\caption{Ablation study evaluating the impact of systematically removing each component from the \method framework, highlighting the contribution of individual techniques to overall performance.}
\label{tab:methods_ablation}
\end{table}

\section{Conclusion}
% \vspace{-2mm}
In this work, we propose a novel universal knowledge distillation framework that effectively addresses the challenges posed by vocabulary mismatches and reasoning-aware distillation. Our \method framework integrates Chain-of-Thought (CoT) augmentation and introduces Cross-CoT Alignment to enhance the transfer of reasoning capabilities from teacher to student models. Furthermore, we extend Optimal Transport beyond token-wise alignment by developing a sequence-level and layer-wise alignment strategy that accommodates varying sequence lengths while preserving contextual integrity.

Extensive experiments demonstrate that \method consistently outperforms existing KD techniques across diverse vocabulary settings, offering improved reasoning capabilities and robustness in domain-specific applications. These results highlight the effectiveness of our approach in bridging gaps in current KD methodologies, making it a more adaptable and reliable solution for real-world model compression. In future work, we aim to extend our approach by incorporating more fine-grained layer-level alignment in the distillation process. This would involve systematically aligning intermediate representations between teacher and student models across multiple layers, allowing for a more comprehensive transfer of knowledge.

\section*{Limitations}
In this study, we hypothesize that the embedding layer, as the initial layer of the model, and the final hidden state, corresponding to the last layer, play a crucial role in knowledge distillation due to their direct influence on representation learning and prediction. However, this assumption may not hold universally across all scenarios, as the contribution of each layer can vary depending on the specific task, model architecture, or training dynamics. Additionally, the role of intermediate layers in the distillation process remains insufficiently explored. A deeper understanding of how knowledge propagates through different layers and how intermediate representations contribute to student model learning could lead to more effective distillation strategies. Future research should investigate adaptive layer-wise distillation techniques that dynamically assign importance to different layers based on task requirements, potentially uncovering new insights into optimal knowledge transfer.

\bibliography{custom}
\bibliographystyle{acl_natbib}

\newpage

\newpage
\appendix
\newpage
\onecolumn
\appendix
\section*{{\huge Appendix}}
\label{sec:appendix}

\section{Sinkhorn Algorithm}

Given $\boldsymbol{\alpha}$, $\boldsymbol{\beta}$ and $\boldsymbol{D}$, the OT distance between empirical probability measures $f$ and $g$ is a linear programing problem:
\begin{equation}
    d_{W}(\boldsymbol{\alpha},\boldsymbol{\beta},\boldsymbol{D}) = \min_{\boldsymbol{T} \in U(\boldsymbol{\alpha},\boldsymbol{\beta})} \langle \boldsymbol{T},\boldsymbol{D} \rangle.
\end{equation}

The solution to obtain the optimal transport plan $\boldsymbol{T}$ is quite computationally expensive. Cuturi \citep{distances2013lightspeed}
introduced an entropy constraint to the transportation polytope, converting the original problem to an entropy regularized optimal transportation problem, resulting in \textit{Sinkhorn distance}, i.e:
\begin{equation}\label{eq5}
\begin{aligned}
    d_{S}^\lambda (\boldsymbol{\alpha}, \boldsymbol{\beta}, \boldsymbol{D}) = \langle \boldsymbol{T}^\lambda, \boldsymbol{D} \rangle 
    \\
    \text{s.t.} \quad \boldsymbol{T}^\lambda = \argmin_{\boldsymbol{T} \in U(\boldsymbol{\alpha}, \boldsymbol{\beta)}} \langle \boldsymbol{T}, \boldsymbol{D} \rangle - \frac{1}{\lambda} h(\boldsymbol{T}),
\end{aligned}
\end{equation}
where $h(\boldsymbol{T}) = - \sum_{i=1}^N \sum_{j=1}^M t_{ij}\log t_{ij}$ is the entropy of $\boldsymbol{T}$. The optimal $\boldsymbol{T}^\lambda$ that minimizes (\ref{eq5}) is:
\begin{equation}
    \label{eq:solution}
    \boldsymbol{T}^\lambda = diag(\boldsymbol{\kappa}_1) \exp^{-\lambda\boldsymbol{D}} diag(\boldsymbol{\kappa}_2)
\end{equation}
where $\exp^{-\lambda\boldsymbol{D}}$ is the element-wise exponential of the matrix $-\lambda\boldsymbol{D}$, $\boldsymbol{\kappa}_1 \in \mathbb{R}^N$, $\boldsymbol{\kappa}_2 \in \mathbb{R}^M$ are the non-negative scaling factors, which can be effectively solved after some Sinkhorn iterations. Hence, the computational cost is greatly reduced compare with the original problem.

\section{Experimental Details}

\paragraph{Models.} We select GPT2-120M \cite{radford2019language},TinyLLaMa-1.1B \cite{zhang2024tinyllama} and GPT2-1.5B as student LLMs. For teachers, we employ GPT2-1.5B, Qwen1.5-1.8B \cite{bai2023qwen}, LLaMa2-7B \cite{touvron2023llama}, Mistral-7B \cite{jiang2023mistral} and Qwen2.5-7B-Instruct \cite{yang2024qwen2}
as teacher LLMs for same and different vocabularies settings. The detail of each models training configurations in KD is listed in Table \ref{tab:kd_configurations}.

\begin{table*}[!ht]
\centering
\begin{adjustbox}{width=0.8\textwidth}
\begin{tabular}{lccc cc ccc}
\toprule
\textbf{Settings} 
    & \multicolumn{3}{c}{\textbf{KD for GPT2-base}} 
    & \multicolumn{2}{c}{\textbf{KD for GPT2}} 
    & \multicolumn{3}{c}{\textbf{KD for TinyLLama}} \\
\cmidrule(lr){2-4} \cmidrule(lr){5-6} \cmidrule(lr){7-9}
    & \textbf{GPT2-base} & \textbf{Qwen1.5} & \textbf{GPT2-large} 
    & \textbf{GPT2-large} & \textbf{Qwen2.5} 
    & \textbf{TinyLLama} & \textbf{LLaMA2} & \textbf{Mistral} \\
\midrule
\textbf{Epoch} 
    & 20 & 10 & 20 
    & 10 & 10 
    & 10 & 10 & 10 \\
\textbf{LR} 
    & $5\times10^{-4}$ & $2\times10^{-5}$ & $5\times10^{-4}$ 
    & $2\times10^{-5}$ & $1\times10^{-3}$ 
    & $1\times10^{-3}$ & $1\times10^{-3}$ & $1\times10^{-3}$ \\
\textbf{Projector LR} 
    & $1\times10^{-3}$ & $1\times10^{-3}$ & $1\times10^{-3}$ 
    & $1\times10^{-3}$ & $1\times10^{-3}$ 
    & $1\times10^{-3}$ & $1\times10^{-3}$ & $1\times10^{-3}$ \\
\textbf{Batch Size} 
    & 32 & 32 & 32 
    & 32 & 32 
    & 32 & 32 & 32 \\
\textbf{LR Scheduler} 
    & Cosine & Cosine & Cosine 
    & Cosine & Cosine 
    & Cosine & Cosine & Cosine \\
\textbf{Fine-Tuning Method} 
    & Full & Full & Full 
    & Full & LoRA 
    & LoRA & LoRA & LoRA \\
\textbf{LoRA Rank} 
    & N/A & N/A & N/A 
    & 256 & 256 
    & 256 & 256 & 256 \\
\textbf{LoRA Alpha} 
    & N/A & N/A & N/A 
    & 8 & 8 
    & 8 & 8 & 8 \\
\textbf{LoRA Dropout} 
    & N/A & N/A & N/A 
    & 0.1 & 0.1 
    & 0.1 & 0.1 & 0.1 \\
\bottomrule
\end{tabular}
\end{adjustbox}
\caption{Detailed training configurations}
\label{tab:kd_configurations}
\end{table*}

\paragraph{Training and Evaluation.}
For GPT2-120M, we employ full fine-tuning for students and teachers, while for TinyLLaMa and GPT2-large we fine-tune the students and teacher with LoRA \cite{hu2021lora}. For evaluation, we sample the responses of models from 5 random seeds. The performance is evaluated by \textbf{ROUGE-L} \cite{lin2004rouge}, a measure of similarity between the generated response and the ground truth. All the experiments are conducted on 4 A100 80GB GPUs.

\paragraph{Detailed Dataset Statistics.} Table \ref{tab:dataset} provides statistics on the number of samples in the training, validation, and test sets for each domain-specific dataset.

% \paragraph{Hyperparameter.} Table \ref{tab:alpha_params} provides detailed hyperparameter $\alpha$ values for each setting.
\paragraph{Hyperparameter.} 

We searched for the hyperaparameter $\alpha$ within the range \(\{0.1, 0.5, 0.6, 0.9\}\) and the best value for each experimental scenario is reported as in Table~\ref{tab:alpha_params}.

% \paragraph{} Table \ref{tab:dataset} provides statistics on the number of samples in the training, validation, and test sets for each domain-specific dataset.

\begin{table}[ht]
\centering
\begin{adjustbox}{width=0.35\textwidth}
\begin{tabular}{l|c|c|c}
\toprule
Dataset   & Train & Validation & Test \\ \midrule
\textbf{Dolly}   & 11435 & 1000 & 500 \\
\textbf{Alpaca}   & 10396 & 500 & 500 \\
\textbf{S-NI}   & 10414 & 500 & 1902 \\
\textbf{DialogSum}   & 12460 & 500 & 1500\\ 
\bottomrule

\end{tabular}
\end{adjustbox}
\caption{Dataset Statistics}
\label{tab:dataset}
\end{table}

\begin{table*}[ht]
\scriptsize
\centering
\begin{adjustbox}{width=0.7\textwidth}
\begin{tabular}{l|c|c|c|c}
\toprule
\textbf{Methods} & \textbf{Dolly} & \textbf{Alpaca} & \textbf{S-NI} & \textbf{Dialogue Sum} \\
\midrule
\multicolumn{5}{l}{\textit{\textbf{Student: GPT2-120M}}} \\
\hline
\multicolumn{5}{c}{\textit{Teacher: GPT2-1.5B (Same Vocabulary)}} \\
\hline
KL + \method & $0.6$ & $0.5$ & $0.5$ & $0.5$ \\
\hline  
\multicolumn{5}{c}{\textit{Teacher: Qwen1.5-1.8B (Different Vocabularies)}} \\
\hline
ULD + \method  & $0.5$ & $0.5$ & $0.5$ & $0.5$ \\
MinED + \method & $0.5$ & $0.5$ & $0.5$ & $0.5$ \\
DSKD + \method  & $0.1$ & $0.9$ & $0.9$ & $0.9$ \\
\bottomrule
\toprule
\multicolumn{5}{l}{\textit{\textbf{Student: TinyLLaMA-1.1B}}} \\
\hline
\multicolumn{5}{c}{\textit{Teacher: Llama2-7B (Same Vocabulary)}} \\
\hline 
KL + \method & $0.6$ & $0.6$ & $0.5$ & $0.5$ \\
\hline
\multicolumn{5}{c}{\textit{Teacher: Mistral-7B (Different Vocabularies)}} \\
\hline 
ULD + \method  & $0.5$ & $0.5$ & $0.6$ & $0.5$ \\
MinED + \method & $0.5$ & $0.9$ & $0.5$ & $0.5$ \\
DSKD + \method  & $0.9$ & $0.9$ & $0.9$ & $0.9$ \\
\bottomrule   
\toprule 
\multicolumn{5}{l}{\textit{\textbf{Student: GPT2-1.5B}}} \\
\hline 
\multicolumn{5}{c}{\textit{Teacher: Qwen2.5-7B-Instruct (Different Vocabularies)}} \\
\hline
ULD + \method  & $0.5$ & $0.5$ & $0.5$ & $0.5$ \\
MinED + \method & $0.5$ & $0.5$ & $0.5$ & $0.9$ \\
DSKD + \method   & $0.9$ & $0.9$ & $0.9$ & $0.9$ \\
\bottomrule
\end{tabular}
\end{adjustbox}
\caption{The best-searched hyperparameters $\alpha$ for different configurations}
\label{tab:alpha_params}
\end{table*}

\label{sec:appendix_exp}

\section{Prompts Details}

\begin{figure*}[ht]
\centering
\begin{tikzpicture}
% Box for Original Prompt
\draw[rounded corners, thick] (0,0) rectangle (7.5,-6);
\node[anchor=north west, text width=7cm] at (0,-0.5) {
\textbf{Original prompt}
\vspace{0.2cm}

Below is an instruction that describes a task, paired with an input that provides further context.\\ 
Write a response that appropriately completes the request.

\texttt{\\\#\#\# Instruction:\{instruction\}\\\#\#\# Input:\{input\}\\\#\#\# Response:}
};

% Box for Zero-shot COT Prompt
\draw[rounded corners, thick] (9,0) rectangle (16.5,-6);
\node[anchor=north west, text width=7cm] at (9,-0.5) {
\textbf{Zero-shot COT prompt}
\vspace{0.2cm}

Below is an instruction that describes a task.\\ 
Write a response with reasoning step by step that appropriately completes the request.

\texttt{\\\#\#\# Instruction:\{instruction\}\\\#\#\# Input:\{input\}\\\#\#\# Response:\\ Let's think step by step.}
};
\end{tikzpicture}
\caption{Original Prompt and Zero-shot COT Prompt}
\label{fig:prompts}
\end{figure*}

\label{sec:prompt}

% \begin{table*}[ht]
% \centering
% \begin{adjustbox}{width=\textwidth}
% \begin{tabular}{l|ccc|cc|ccc}
% \toprule
% \parbox[c][2.5em][c]{1.5cm}{\centering \textbf{Settings}}
%     & \multicolumn{3}{c|}{\textbf{KD for GPT2-base}}
%     & \multicolumn{2}{c|}{\textbf{KD for GPT2}}
%     & \multicolumn{3}{c}{\textbf{KD for TinyLLama}} \\
% \cmidrule{2-9}
% & \textbf{GPT2-base} & \textbf{Qwen1.5} & \textbf{GPT2-large}
% & \textbf{GPT2-large} & \textbf{Qwen2.5}
% & \textbf{TinyLLama} & \textbf{LLaMA2} & \textbf{Mistral} \\
% \midrule
% \textbf{Epoch}
%     & 20 & 10 & 20 
%     & 10 & 10
%     & 10 & 10 & 10 \\
% \textbf{LR}
%     & 5e-4 & 2e-5 & 5e-4
%     & 2e-5 & 1e-3
%     & 1e-3 & 1e-3 & 1e-3 \\
% \textbf{Projector LR}
%     & 1e-3 & 1e-3 & 1e-3
%     & 1e-3 & 1e-3
%     & 1e-3 & 1e-3 & 1e-3 \\
% \textbf{Batch Size}
%     & 32 & 32 & 32
%     & 32 & 32
%     & 32 & 32 & 32 \\
% \textbf{LR Scheduler}
%     & Cosine & Cosine & Cosine
%     & Cosine & Cosine
%     & Cosine & Cosine & Cosine \\
% \textbf{Fine-Tuning Method}
%     & Full & Full & Full
%     & Full & LoRA
%     & LoRA & LoRA & LoRA \\
% \textbf{LoRA Rank}
%     & N/A & N/A & N/A
%     & 256 & 256
%     & 256 & 256 & 256 \\
% \textbf{LoRA Alpha}
%     & N/A & N/A & N/A
%     & 8 & 8
%     & 8 & 8 & 8 \\
% \textbf{LoRA Dropout}
%     & N/A & N/A & N/A
%     & 0.1 & 0.1
%     & 0.1 & 0.1 & 0.1 \\
% \bottomrule
% \end{tabular}%
% \end{adjustbox}
% \caption{Detailed training configurations}
% \label{tab:kd_configurations}
% \end{table*}

\end{document}